\documentclass[conference]{IEEEtran}
\IEEEoverridecommandlockouts

\usepackage{cite}
\usepackage{url}
\usepackage{amsmath,amssymb,amsfonts}

\usepackage[font=footnotesize,labelfont=bf]{caption}
\usepackage{subcaption}

\newtheorem{theorem}{Theorem}
\usepackage{algorithmic}
\usepackage{balance}
\usepackage{graphicx}
\usepackage{textcomp}
\usepackage{xcolor}
\usepackage{booktabs}
\usepackage{multirow}
\usepackage{geometry}
\usepackage{longtable}
\usepackage{lscape}
\def\BibTeX{{\rm B\kern-.05em{\sc i\kern-.025em b}\kern-.08em
    T\kern-.1667em\lower.7ex\hbox{E}\kern-.125emX}}
\newcommand{\email}[1]{#1}
\newcommand{\spacetune}[1]{#1}
\begin{document}

\title{Well-Conditioned Polynomial Representations
for Mathematical Handwriting Recognition
}

\author{\IEEEauthorblockN{Robert M. Corless }
\IEEEauthorblockA{\textit{Department of Computer Science,} \\
\textit{Western University, }\\
London, Canada\\
\email{rcorless@uwo.ca}}
\and
\IEEEauthorblockN{Deepak Singh Kalhan}
\IEEEauthorblockA{\textit{Cheriton School of Computer Science,} \\
\textit{University of Waterloo, }\\
Waterloo, Canada\\
 \email{dsinghkalhan@uwaterloo.ca}
}
\and
\IEEEauthorblockN{Stephen M. Watt}
\IEEEauthorblockA{\textit{Cheriton School of Computer Science,} \\
\textit{University of Waterloo, }\\
Waterloo, Canada\\
 \email{smwatt@uwaterloo.ca}
}

}

\maketitle
\begin{abstract}
Previous work has made use of a parameterized plane curve polynomial representation for mathematical handwriting, with the polynomials represented in a Legendre or Legendre--Sobolev graded basis.  
This provides a compact geometric representation for the digital ink. 
Preliminary results have also been shown for Chebyshev and Chebyshev--Sobolev bases.   
This article explores the trade-offs between basis choice and polynomial degree to achieve accurate modeling with a low computational cost.
To do this, we consider the condition number for polynomial evaluation in these bases and bound how the various inner products give norms for the variations between symbols.
\end{abstract}


\section{Introduction}
Digital ink refers to the representation and manipulation of ink strokes on a digital surface, allowing users to write, draw, and create content digitally. These traces are often captured as a sequence of $(x, y)$ coordinates over time and are used in applications such as note-taking, signature verification to handwriting, and mathematical expression recognition. To make effective use of digital ink, it is important to represent these traces in a form that is both compact and suitable for analysis or recognition~\cite{mohamed2009concepts}.

One approach is to approximate the traces using parameterized plane curves defined by polynomials~\cite{CharWatt,golubitsky2010distance,watt2012polynomial}. The $x$ coordinates and $y$ coordinates are each written as functions of arc length, and these functions are approximated by polynomials. Choosing how to represent these polynomials has a significant impact on both the accuracy of the model and the cost of computation.

The most familiar representation of polynomials uses the monomial basis: ($1, s, s^2, \dots$). However, this basis can be ill-conditioned, especially when working with higher-degree polynomials. The use of orthogonal polynomials such as Legendre, Laguerre, Chebyshev or other polynomials provides a powerful framework for approximating functions, which can frequently be better-conditioned than the monomial basis.

Questions on conditioning of polynomial bases can be subtle.  See~\cite{farouki1987numerical} for a coherent mathematical framework and~\cite{farouki1996optimal} for a proof that the Bernstein polynomial basis used in Computer-Aided Geometric Design is \textsl{optimal} over all bases nonnegative on an interval.  See~\cite{corless2004bernstein} for a proof that Lagrange bases are similarly optimal over all bases nonnegative on a finite set of points.  Another view of that same fact can be found in~\cite{CARNICER201952}.  

The key point of these analyses is that different bases have different condition numbers for the same polynomial, and that difference can be exponentially large.  Another important point is that condition numbers reflect sensitivity not just to errors in computation, but errors in the original data: an ill-conditioned representation will result in large changes in the coefficients when small changes in the data are seen. This directly impacts character recognition rates.  

It is important to choose a suitable basis at the start of computation, because changing bases during the computation also carries a risk of numerical instability because the change-of-basis matrices are frequently ill-conditioned, with (ordinary) condition numbers that are exponential in the degree.  Working with such matrices requires special care in floating-point arithmetic.

Ill-conditioned polynomials are \textsl{also} sensitive to rounding errors in floating-point arithmetic, but recent work using bidiagonal factoring of the totally nonnegative matrices involved, e.g.~\cite{marco2024total}, can mitigate that greatly.
Use of bidiagonal factoring can also reduce numerical instability if the polynomial basis is changed during the computation.  The sensitivity to data errors remains, however.

The Legendre basis consists of polynomials that are orthogonal with respect to the standard inner product in the interval $[ -1, 1 ]$. This orthogonality makes the Legendre basis a good choice for approximating smooth functions and for minimizing the overall error in representation. The Legendre--Sobolev basis also takes into account not only the function values but also the derivatives, leading to a basis that better reflects the geometric features of a curve such as curvature and slope changes. This series has gained attention for its improved recognition rates \cite{mazalov2012improving}. 

Some recent work has explored the use of Chebyshev \cite{watt2012polynomial} and Chebyshev--Sobolev bases \cite{kalhan2024first}. Chebyshev polynomials are orthogonal and provide good numerical stability \cite{wu2015new}, but their weight function differs from the Legendre basis, giving them different approximation characteristics \cite{lanczos1952solution}. The Chebyshev--Sobolev basis also incorporates derivative terms and further enhances modeling accuracy for certain applications. In \cite{kalhan2024first}, the suitability of the Chebyshev--Sobolev series for handwritten text and symbol recognition was first explored. The present work is an extension of this line of research with a comparison of Chebyshev-Sobolev bases to Legendre, Legendre-Sobolev and Chebyshev bases, with the aim of identifying the effective trade-offs between accuracy and computational cost.

Although various polynomial basis exist, it is not always clear to make the optimal choice. Each basis has different trade-offs between accuracy, and computational cost. A highly accurate basis might be computationally expensive whereas, a simpler basis may be easier but not accurate. In this work we compare different polynomial basis to make optimal selection considering accuracy and computational time.

A key tool in our analysis is the condition number associated with polynomial evaluation in a given basis. The condition number measures the sensitivity of the output of a function to changes in its input \cite{shetty2007handwritten}. In the context of polynomial evaluation, a high condition number implies that small changes in inputs can result in large changes in output — which is undesirable for handwriting applications that require precision and robustness. 

In this work, we have also derived a bound on the Sobolev norm of the difference between two polynomials, showing that it can be controlled by the infinity norm of the difference in their coefficient vectors. Specifically, we bound the expression \( ||f(x) - g(x)||_s \) in terms of \( ||\mathbf{f} - \mathbf{g}||_\infty \), scaled by factors involving the differentiation matrix and the basis norm. This provides a practical measure of stability, ensuring that small changes in coefficients lead to proportionally small changes in both the function shape and its derivative, an essential property for accurate and robust handwriting modeling.

In summary, our goal is to provide clear guidance on selecting the most suitable basis and polynomial degree for practical handwriting modeling. We do this through both theoretical analysis and evaluation on a real dataset of mathematical symbols. Our results aim to improve the efficiency and reliability of digital ink analysis, with potential applications in education, accessibility, and user-friendly digital interfaces.

\section{Parametric Curve Representation}

Representing digital ink as a parametric curve is a new direction compared to older methods that used pixels or tracked points over time. Those earlier methods are highly dependent on resolution - in space or time - which can create problems with precision \cite{mazalov2012improving} and make the results less reliable.

In Parametric Curve Representation, both the $x$ and $y$ positions of the writing tip are expressed as functions of a single parameter, typically denoted as $s$. This parameter is arc length s,  a normalized value in the range $[-1,1]$ indicating progression along the stroke, where
\begin{align}
    ds^2 = dx^2 + dy^2.
\end{align}
We define a handwriting stroke using two functions:
\[
x(s), \quad y(s), \quad s \in [-1, 1].
\]
These functions specify the horizontal and vertical positions of the pen w.r.t normalized arc length. This creates a smooth 2D path that models the stroke as a continuous curve.

Parametric curves are effective because handwriting involves smooth and continuous motion. Rather than handling a large sequence of discrete points, we represent the entire stroke using functions.

Given a functional inner product, a graded basis of orthogonal polynomials $\{B_i(s)\}_{i=0,....d}$ may be obtained using Gram-Schmidt orthogonalization of the set of monomials $\{s^i\}_{i \in [0..d]}$. Now, using $\{B_i(s)\}_{i=0,....d}$ for approximating ($x(s)$, $y(s)$), a trace can be represented as 
\begin{align}
    &x(s) \approx \sum_{i=0}^{d} x_i B_i(s)  \\
    &y(s) \approx \sum_{i=0}^{d} y_i B_i(s)
\end{align}
where, $x_i$ and $y_i$ are the coefficients and $d$ is the degree of the truncated series. 
The coefficients $x_i$ and $y_i$ determine the shape and direction of the curve. By choosing appropriate values for these coefficients, we can match the curve very close to the actual pen trajectory.

In this method, we store only the function’s coefficients instead of all the stroke points. This makes the representation more compact and flexible. The quality of the result depends on two things: the degree of the polynomial and the type of basis used. We will show that some bases give better accuracy even at low degrees, while others may perform worse as the degree increases. In this work, we use parametric polynomial curves to model handwriting strokes. In later sections, we look at how different bases affect the accuracy, stability, and efficiency of this approach.

\section{Choice of Basis}
\label{sec:basis}

When we represent handwritten symbols using parameterized polynomial curves, the choice of basis for these polynomials greatly impacts how well the representation performs. A basis is a set of building blocks from which we construct the polynomial. Standard bases include monomials, Legendre polynomials, Chebyshev polynomials, and their Sobolev variants. Each basis has its mathematical properties, and some are better suited than others for modeling handwriting stably and efficiently.

\subsection*{Functional Inner Product and Orthogonality}

Understanding why some bases work better than others helps to look at the idea of a \textit{functional inner product}. Given two functions $f(s)$ and $g(s)$, the standard inner product over the interval $[-1, 1]$ is defined as
\begin{align}
\langle f, g \rangle = \int_{-1}^{1} f(s) g(s) \, ds.
\end{align}
Using the Gram-Schmidt process, one can generate the basis polynomials.
This property ensures that each function captures independent information, which reduces redundancy and improves numerical stability when fitting curves to data.

Orthogonal bases like the Legendre and Chebyshev polynomials are defined using functional inner product with weights 1 and $\frac{1}{\sqrt{1-x^2}}$ respectively. Because of orthogonality, there is no interfence in the calculation of coefficients. They also help avoid problems like overfitting or numerical instability when using higher-degree polynomials \cite{bani2013orthogonal}.

\subsection*{Sobolev Norms: Incorporating Derivatives}

The standard inner product only considers the values of functions, sometimes it is essential to include their derivatives. In handwriting, the slope and curvature of a stroke can be just as important as its position. To account for this, we use a \textit{Sobolev inner product}, which includes derivatives in the definition:
\begin{align}
\langle f, g \rangle_{S} = \int_a^b f(s) g(s) \, ds + \mu \int_a^b f'(s) g'(s) \, ds,
\end{align}
where $\mu$ is a real parameter that controls how much emphasis is placed on the derivative terms and we assume its fixed. This inner product leads to what we call \textit{Sobolev orthogonal polynomials}.

Using a Sobolev norm leads to smoother polynomial curves that not only match the position of the handwriting but also its velocity and shape characteristics. This approach can lead to better modeling of subtle differences between symbols, which is valuable for recognition and classification tasks \cite{golubitsky2010distance}.

\subsection*{Why Does This Matter?}

The selection between a standard orthogonal basis and a Sobolev basis has a significant affect on representing handwriting strokes. A standard basis may fit the overall shape of the stroke well but may overlook how sharply it bends or how smoothly it flows. A Sobolev basis captures these details, which can be helpful in distinguishing visually similar characters or symbols \cite{golubitsky2010distance}.

The choice of inner product affects the norm used to measure approximation error. The standard $L^2$ norm measures the square error at position only, while the Sobolev norm combines position error with derivative error. Choosing norms that better reflect the visual and geometric features of handwriting can enhance both accuracy and robustness \cite{golubitsky2010distance}.

In this work, we compare different bases Legendre, Legendre--Sobolev, Chebyshev, and Chebyshev--Sobolev to understand the influence of basis choice on model performance. Our aim is to find representations that balance accuracy, and computation cost.  In what follows the degrees of the polynomials are $d$ and the length of the coefficient vectors is $n = d+1$.

\begin{theorem}
Let \( f(x) \) and \( g(x) \) be polynomials expressed in an orthogonal basis \( \{P_i(x)\} \), with coefficient vectors \( \mathbf{f} \) and \( \mathbf{g} \), respectively. Suppose the polynomials and their derivatives are evaluated on the interval \([0,1]\), and let \( \mathbf{D} \) be the differentiation matrix corresponding to the basis. Then, the Sobolev norm of the difference between \( f \) and \( g \) satisfies
\[
\|f - g\|_s \leq \sqrt n \; \|\mathbf{f} - \mathbf{g}\|_\infty (1 + \mu \|\mathbf{D}\|),
\]
where \( \mu \) is the weight for the derivative term.
\end{theorem}

\begin{IEEEproof}
    Let $f(x)$ and $g(x)$ be polynomials given in orthogonal basis $P_i(x)$ that
 \begin{align}
     f(x) &= \sum_{i=0}^d f_i P_i(x) & g(x) &= \sum_{i=0}^d g_i P_i(x)\\
     &= \mathbf{f} \mathbf{P} &&= \mathbf{g} \mathbf{P},
 \end{align}
where, $\mathbf{f}$ is a vector of polynomial coefficients and $\mathbf{P}$ is a vector of orthogonal basis.

Considering section 11.2, 
in \cite{corless2013graduate}, there exists a differentiation matrices $\mathbf{D}$ such that
 \begin{align}
     \mathbf{f}' &= \mathbf{D} \mathbf{f}& \mathbf{g}' &= \mathbf{D} \mathbf{g},\\
      \mathbf{f}' \mathbf{P} &= \mathbf{D} \mathbf{f} \mathbf{P} & \mathbf{g}' \mathbf{P} &= \mathbf{D} \mathbf{g} \mathbf{P}\\
       \mathbf{f}'(x) &= \mathbf{D} \mathbf{f}(x)& \mathbf{g}'(x) &= \mathbf{D} \mathbf{g}(x)    \end{align}
 where $f'$ is a vector of coefficients for the
derivative of $f(x)$ in the same basis. Now,
\begin{align}
||f(x)-g(x)||_s &= ||f(x)-g(x)|| +\mu ||f'(x)-g'(x)||\\
&= ||f(x)-g(x)|| + \mu ||\mathbf{D}\mathbf{f}(x)-\mathbf{D}\mathbf{g}(x) ||
\end{align}
Using Cauchy-Schwartz inequality, we can expand the terms as
\begin{align}
||f(x)-g(x)||_s \leq{}&  ||\mathbf{f}(x)-\mathbf{g}(x)  || \hspace{0.1 cm} \nonumber \\&{}+ \mu ||\mathbf{D}||\hspace{0.1 cm}||\mathbf{f}(x)-\mathbf{g}(x)|| \hspace{0.1 cm}\\
={}&||\mathbf{f}(x)-\mathbf{g}(x)  || \hspace{0.1 cm} (1+\mu \hspace{0.1 cm} ||D||)
\end{align}

We know that $||f(x)-g(x)||_2 \leq \sqrt n \;\hspace{0.1 cm}||f(x)-g(x)||_{\infty}$. Here, $\mathbf{f}(x)$ and $\mathbf{g}(x)$ are defined in (0,1).

\begin{align}
||f(x)-g(x)||_s\leq  \sqrt n\; ||\mathbf{f}(x)-\mathbf{g}(x)  ||_{\infty} \hspace{0.1 cm} (1+\mu \hspace{0.1 cm} ||D||)
\end{align}
\end{IEEEproof}

\begin{theorem}
Let \( f(x) \) and \( g(x) \) be polynomials expressed in an orthogonal basis \( \{P_i(x)\} \), with coefficient vectors \( \mathbf{f} \) and \( \mathbf{g} \). Let \( \mathbf{D} \) be the differentiation matrix such that \( \mathbf{f}' = \mathbf{D} \mathbf{f} \) and \( \mathbf{g}' = \mathbf{D} \mathbf{g} \), and assume \( f(x) \), \( g(x) \), and their derivatives are evaluated over an interval where the basis functions are bounded. Then, the Sobolev norm of the difference between \( f \) and \( g \) satisfies:
\[
\|f(x) - g(x)\|_s \leq \sqrt{n} \cdot (1 + \mu \|\mathbf{D}\|) \cdot \|\mathbf{P}\| \cdot \|\mathbf{f} - \mathbf{g}\|_\infty,
\]
where  \( \mu \) is the weight in the Sobolev norm, and \( \|\mathbf{P}\| \) is the norm of the basis vector evaluated over the domain.
\end{theorem}
\begin{IEEEproof}
Let $f(x)$ and $g(x)$ be polynomials given in orthogonal basis $P_i(x)$ that
 \begin{align}
     f(x) &= \sum_{i=0}^d f_i P_i(x) & g(x) &= \sum_{i=0}^d g_i P_i(x)\\
     &= \mathbf{f} \mathbf{P} &&= \mathbf{g} \mathbf{P},
 \end{align}
where, $\mathbf{f}$ is a vector of polynomial coefficients and $\mathbf{P}$ is a vector of orthogonal basis.

Considering section 11.2, 
in \cite{corless2013graduate}, there exists a differentiation matrices $\mathbf{D}$ such that
 \begin{align}
     \mathbf{f}' &= \mathbf{D} \mathbf{f}& \mathbf{g}' &= \mathbf{D} \mathbf{g},
 \end{align}
 where $f'$ is a vector of coefficients for the
derivative of $f(x)$ in the same basis. Now,
\begin{align}
||f(x)-g(x)||_s &= ||f(x)-g(x)|| +\mu ||f'(x)-g'(x)||\nonumber\\
&= ||(\mathbf{f}-\mathbf{g}) \mathbf{P} || + \mu ||(\mathbf{D}\mathbf{f}-\mathbf{D}\mathbf{g}) \mathbf{P} ||
\end{align}
Using the Cauchy--Schwartz inequality, we can expand the terms as
\begin{align}\label{norm_coeff}
||f(x)-g(x)||_s &\leq  \nonumber\\&||(\mathbf{f}-\mathbf{g})  || \hspace{0.1 cm}||\mathbf{P}|| + \mu ||\mathbf{D}||\hspace{0.1 cm}||(\mathbf{f}-\mathbf{g})|| \hspace{0.1 cm}||\mathbf{P}||
\end{align}

We know that  $||f-g||_2 \leq \sqrt{n}\hspace{0.1 cm}||f-g||_{\infty}$. Here, $\mathbf{f}$ and $\mathbf{g}$ are n dimension vectors.

\begin{align}
||f(x)-g(x)||_s &\leq  \sqrt{n}\hspace{0.1 cm} ||(\mathbf{f}-\mathbf{g})  ||_{\infty} \hspace{0.1 cm}||\mathbf{P}|| \nonumber \\  + &\sqrt{n}\mu \hspace{0.1 cm}||\mathbf{D}||\hspace{0.1 cm}||(\mathbf{f}-\mathbf{g})||_{\infty}\hspace{0.1 cm}||\mathbf{P}||\nonumber\\
= &\sqrt{n} \hspace{0.2 cm}[\hspace{0.1 cm}1 \hspace{0.1 cm} + \mu \hspace{0.1 cm}||\mathbf{D}||\hspace{0.1 cm}]\hspace{0.2 cm}||\mathbf{P}||\hspace{0.1 cm} ||(\mathbf{f}-\mathbf{g})||_{\infty}\
\end{align}
    
\end{IEEEproof}

\section{Experiments}
\subsection{Experimental setting}
For our experiments, we have run tests on UCI pendigits dataset \cite{misc_pen-based_recognition_of_handwritten_digits_81} and Mathwriting dataset \cite{gervais2025mathwriting} . The UCI pendigits dataset contains samples of 10992 handwritten digits (0-9) of multiple users. For each symbol, the number of strokes and the $x$ and $y$ coordinates of the sample
points are available.  
The normalized Legendre--Sobolev coefficient vectors and 
Chebyshev--Sobolev coefficient vectors were computed for all
samples, with the value of the parameter $\mu$ set to 1/8. The reason for the choice of this value is proven results in \cite{kalhan2024first} and \cite{golubitsky2010distance}. 
\subsection{Coefficient Norms Across Polynomial Bases}
For each basis (Chebyshev, Legendre, Legendre--Sobolev, and Chebyshev--Sobolev), we generate the sequences of orthogonal polynomials up to a fixed degree. We then compute the coefficients for the orthogonal vectors for the given handwritten mathematical symbols. 

\begin{figure}[t]
    \centering
    \includegraphics[width=0.48\textwidth,height=6cm]{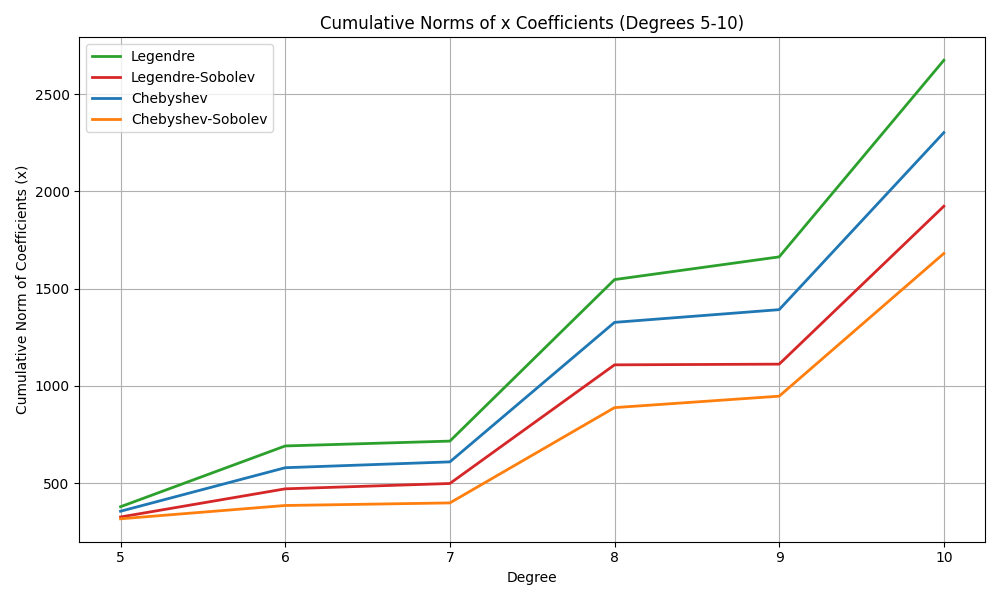}
    \caption{Cumulative norms for degrees 5--10.}
    \label{fig:cunorms1}
\end{figure}

As seen in \eqref{norm_coeff}, the Sobolev norm is bounded by the norm of coefficient vector, we analyze the norm of coefficient vector considering series up to degree 20.

A lower value of norm of coefficient vector implies that the basis can represent the function compactly and efficiently, while a growing norm with increasing degree may indicate poor efficiency. We compare these trends across the bases to highlight stability and efficiency of basis for approximating real handwriting data.

\begin{figure}[t]
    \centering
    \includegraphics[width=0.48\textwidth,height=6cm]{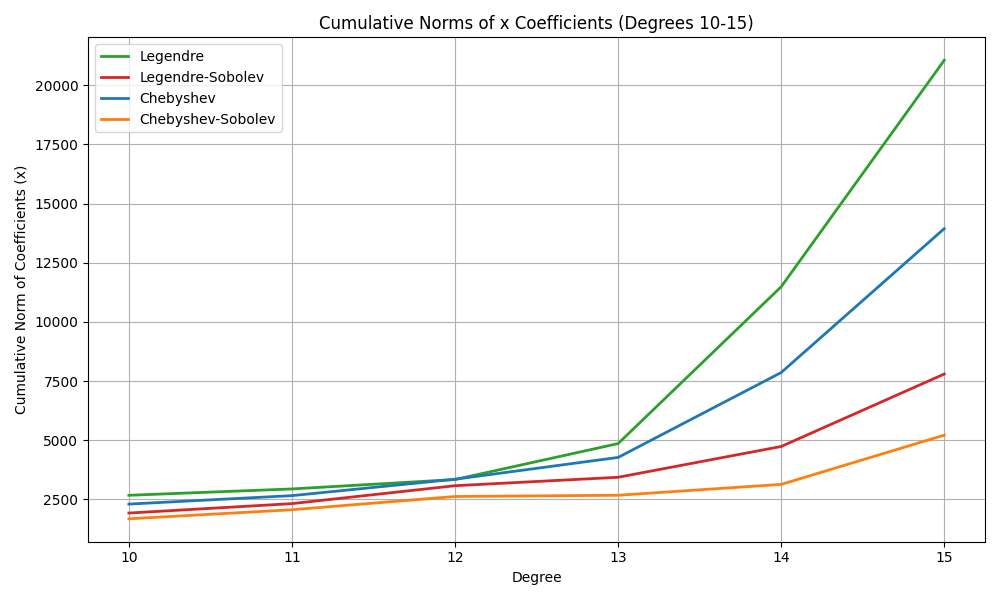}
    \caption{Cumulative norms for degrees 10--15.}
    \label{fig:cunorms2}
\end{figure}

\begin{figure}[t]
    \centering
    \includegraphics[width=0.48\textwidth,height=6cm]{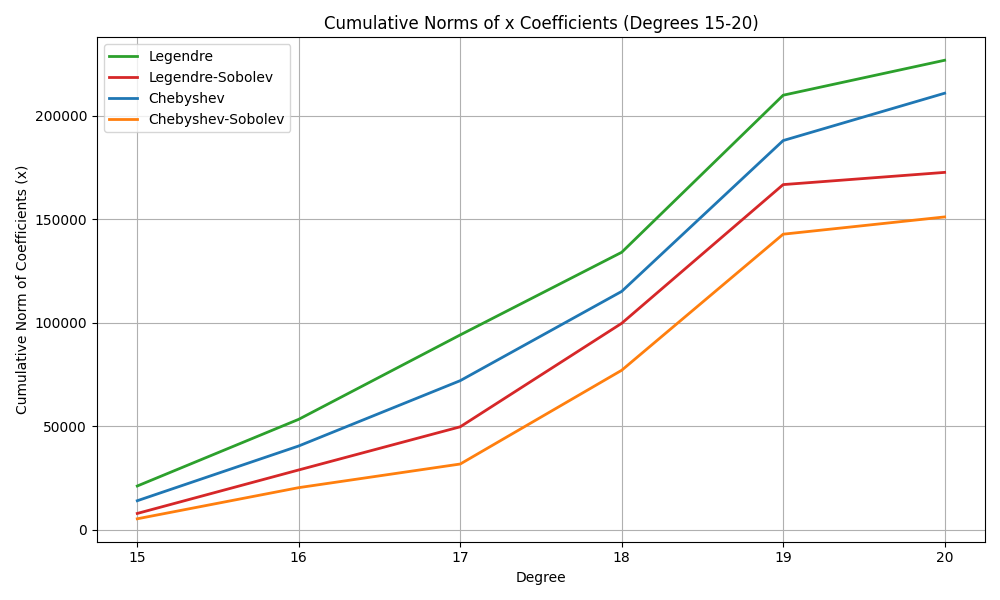}
    \caption{Cumulative norms for degrees 15--20.}
    \label{fig:cunorms3}
\end{figure}

Figures \ref{fig:cunorms1}, \ref{fig:cunorms2}, and \ref{fig:cunorms3} show the cumulative norms of polynomial coefficients considering up to degrees 10, 15 and 20, respectively.
In Figure \ref{fig:cunorms1}, the cumulative norms of coefficients across degrees 5–10 show apparent differences between the four polynomial bases. At degree 5, all the polynomial bases start with modest values, showing a slight difference at lower degrees. Legendre and Chebyshev bases show steeper growth as the degree increases, with Legendre consistently producing the largest cumulative norms, particularly at degrees 8–10. Chebyshev follows a comparable but slightly more restrained trajectory. In contrast, both Legendre–Sobolev and Chebyshev-–Sobolev maintain more reasonable increases, illustrating the stabilizing effect of the Sobolev inner product. This controlled change indicates that the derivative-sensitive Sobolev bases overcome coefficient explosion, leading to smooth approximations at higher degrees. By degree 10, the gap is high. Legendre dominates all the bases, Chebyshev is slightly lower, whereas the Sobolev variants remain compact, highlighting their advantage for numerical stability.


As shown in Figure \ref{fig:cunorms2}, from degrees 10–15, the differences between the bases become more evident. The Legendre basis grows the fastest, especially after degree 13, while Chebyshev also grows quickly but stays slightly lower. The Sobolev-based bases increase much more slowly. Legendre–Sobolev shows stable but smaller growth, and Chebyshev–Sobolev remains the most sturdy with the least increase. This indicates that the Sobolev bases manage the coefficient growth and give smoother results for higher degrees.

Figure \ref{fig:cunorms3} highlights the cumulative norm of coefficients growth from degree 15 to 20. The gap between the bases becomes significantly large. Legendre and Chebyshev bases rise sharply, making them less trustworthy for higher degrees. The Sobolev versions stay much controlled. Legendre–Sobolev grows steadily, and Chebyshev–Sobolev again has the gradual and most stable increase. Sobolev bases, especially Chebyshev–Sobolev, give smoother and more stable results for high-degree handwriting representation.

\subsection{Computation Time Analysis of Polynomial Bases}

\begin{figure}[t]
    \centering
    \includegraphics[width=0.45\textwidth,height=6cm]{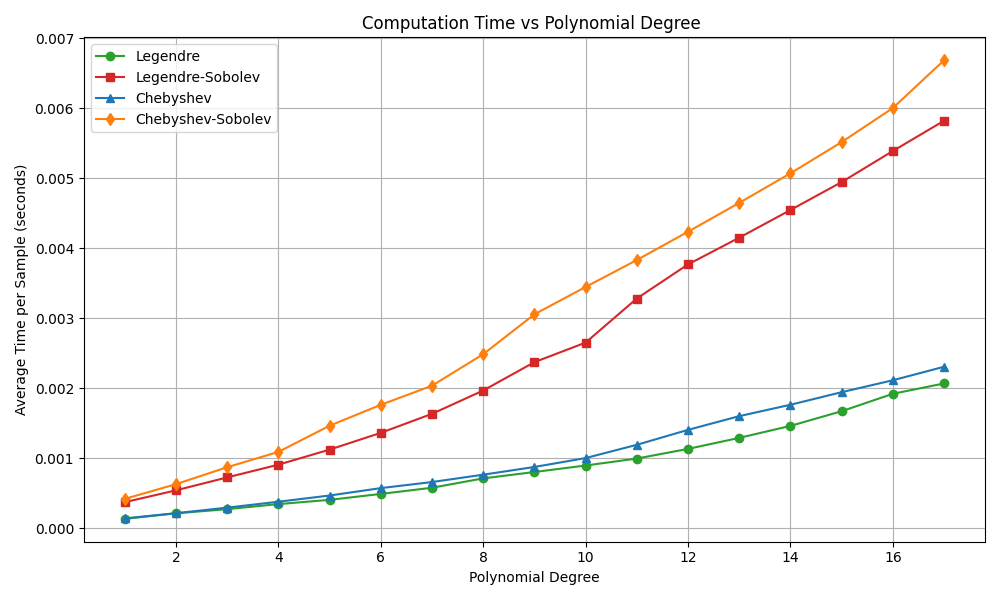}
    \caption{Computation time per sample as a function of polynomial degree 
             for different polynomial bases (Legendre, Legendre-Sobolev, 
             Chebyshev, Chebyshev-Sobolev).}
    \label{fig:comp-time}
\end{figure}

In this experiment, we compare the computation time to represent handwritten ink symbols using polynomial bases for different degrees. We have considered Legendre, Chebyshev, Legendre-Sobolev and Chebyshev--Sobolev basis. 
The parameterized curve for each symbol is projected onto the considered polynomial basis. This involved the computation of inner products and applying numerical integration to calculate coefficients using Gram-Schmidt orthogonalized bases. The experiment was repeated for different degrees and the recorded computation time was averaged over the entire dataset.

\begin{figure*}[t]
    \centering
    \includegraphics[width=0.95\textwidth]{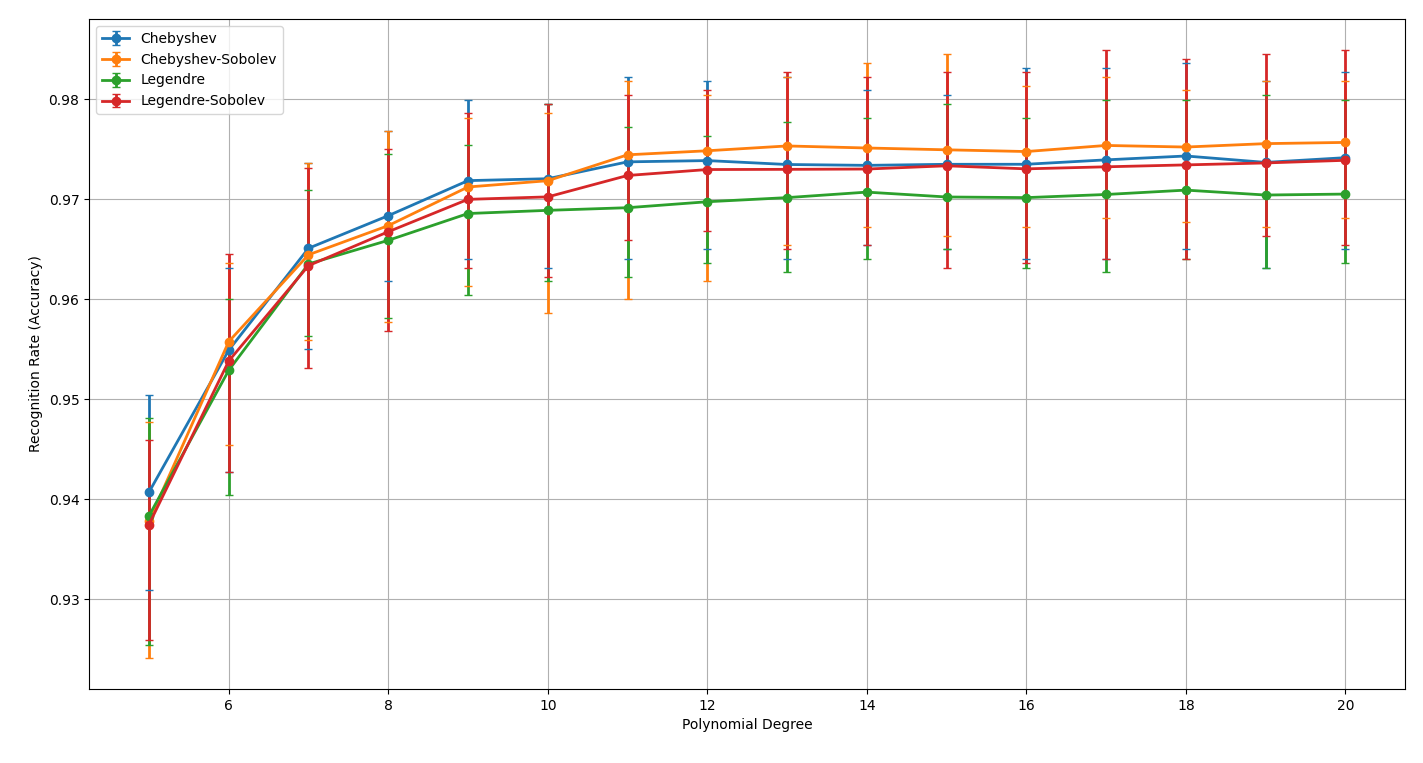}
    \spacetune{\vspace{-0.5\baselineskip}}
    \caption{Recognition rate (accuracy) as a function of polynomial degree for different polynomial bases, with error bars showing variability.}
    \label{fig:reco-rate}
\end{figure*}
\begin{figure*}[thbp]
    \centering
    \begin{subfigure}{0.95\textwidth}
        \centering
        \includegraphics[height=0.166\textheight]{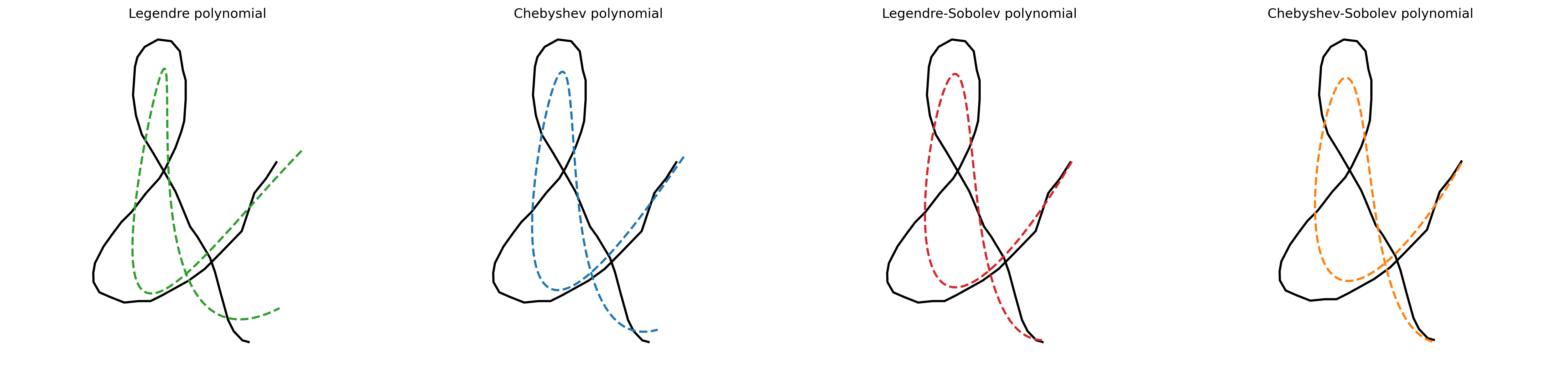}
        \caption{Approximation at degree 5}
        \label{fig:trace5}
    \end{subfigure}

    \begin{subfigure}{0.95\textwidth}
        \centering
        \includegraphics[height=0.166\textheight]{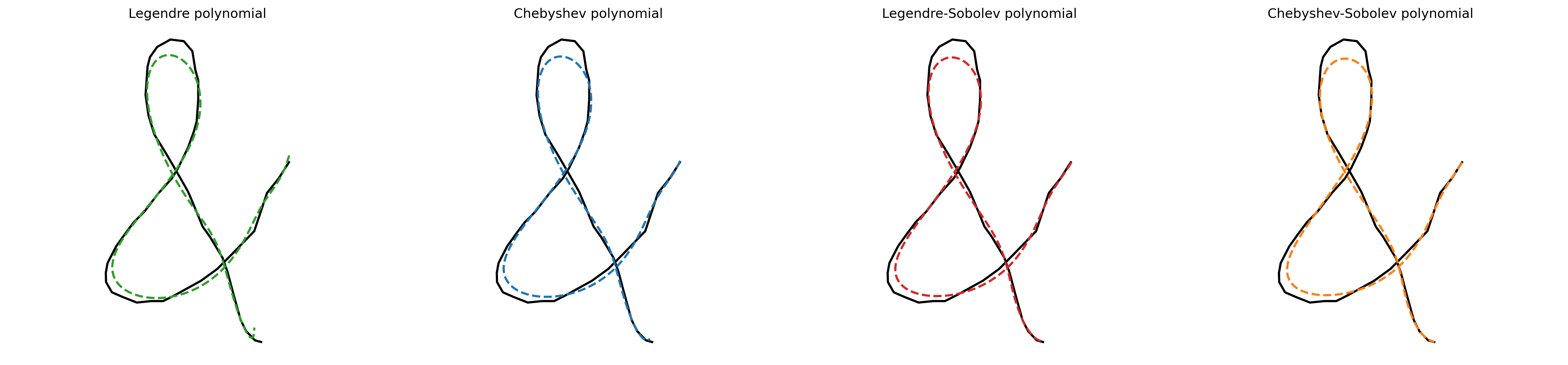}
        \caption{Approximation at degree 10}
        \label{fig:trace10}
    \end{subfigure}

    \begin{subfigure}{0.95\textwidth}
        \centering
        \includegraphics[height=0.166\textheight]{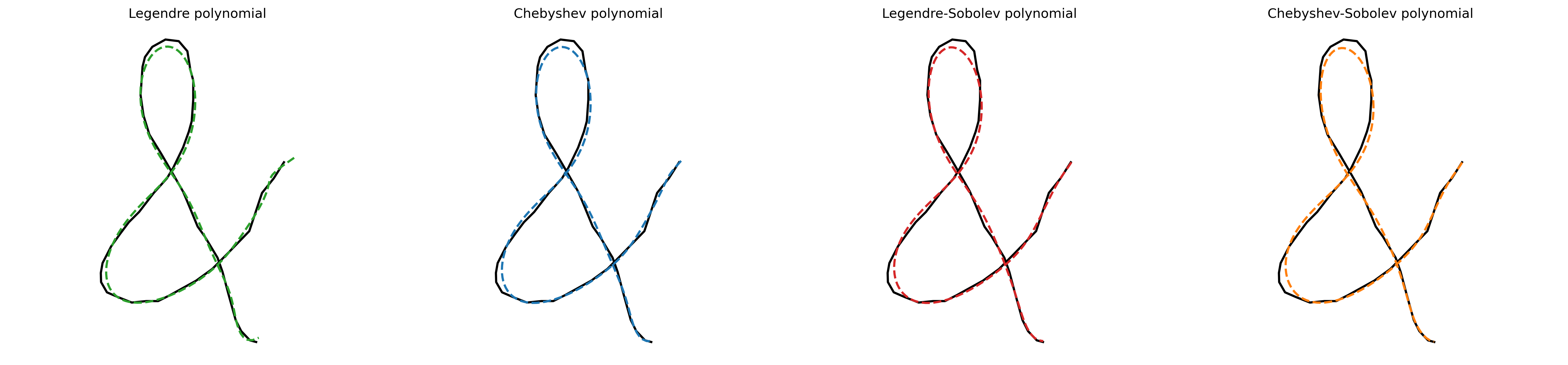}
        \caption{Approximation at degree 15}
        \label{fig:trace15}
    \end{subfigure}

    \begin{subfigure}{0.95\textwidth}
        \centering
        \includegraphics[height=0.166\textheight]{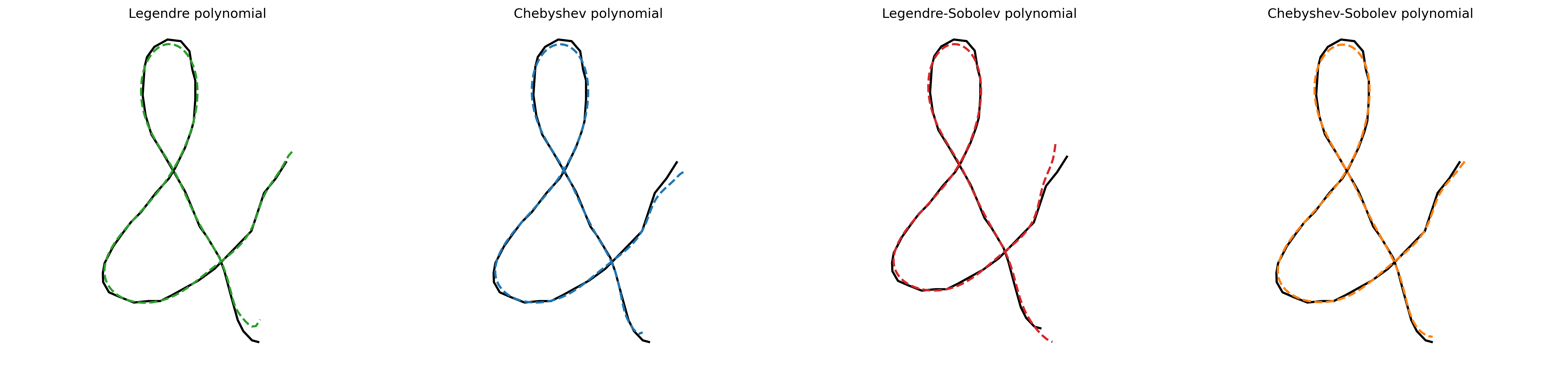}
        \caption{Approximation at degree 20}
        \label{fig:trace20}
    \end{subfigure}

    \caption{
    Polynomial approximations of a representative symbol using different bases across degrees 5, 10, 15, and 20. \newline The solid black line show piecewise-linear approximation of the original data, while the dotted colored lines represent polynomial approximations.
    }
    \label{fig:trace-all}
\end{figure*}

Figure \ref{fig:comp-time} highlights the average computation time per sample of different polynomial bases varying with degree. The time grows almost linearly with degree, which is expected since the number of inner products also increases linearly. Legendre and Chebyshev bases have nearly similar timings. Still, Chebyshev is slightly slower because its inner product involves the weight factor $\tfrac{1}{\sqrt{1-x^2}}$, unlike Legendre, which has a constant weight of $1$. The Sobolev bases require more time because of the inclusion of derivative terms in the inner product. Among them, Chebyshev–Sobolev is the most expensive, as it combines the derivative term with the $\tfrac{1}{\sqrt{1-x^2}}$ weight, making the computations expensive. There are small fluctuations at some degrees; however, they are insignificant.

\subsection{Recognition Rate}
We evaluate the recognition rate using the UNIPEN dataset, focusing on handwritten digits from 0 to 9. Each digit was preprocessed by uniformly resampling its trajectory to exactly 8 points, ensuring consistent representation across all samples. These 8-point traces are then represented as a parametric curve representation, and the resulting polynomial coefficients are obtained as feature vectors. This representation captures the essential shape and dynamics of the handwritten digits while enabling compact representation for classification.

We use a support vector machine (SVM) framework for the classification task. The dataset is divided as 80 \% for training and 20 \% for testing. A one-vs-one classification strategy is adopted, which results in 45 binary SVM classifiers, each trained to distinguish between a pair of 10-digit classes. During inference, each test sample goes through all 45 classifiers. The final predicted label is determined using majority voting based on the outputs of these binary classifiers. This approach allows robust multi-class recognition while leveraging the discriminative power of binary SVM decision boundaries.

We randomly split the dataset into training and testing sets 100 times to obtain statistically trustworthy results. Recognition accuracy is computed for each split, and the reported results are based on these repeated trials. We further summarize the outcomes at each polynomial degree by calculating the maximum, minimum, and mean recognition rates across the 100 runs for all polynomial bases considered.

Figure \ref{fig:reco-rate} shows the recognition rates of different polynomial bases as the degree varies from 5 to 20. The curves are unsmooth due to the natural variations across repeated runs. Performance improves steadily up to around degree 10–12, after which the gains become gradual and the curves almost level off. Chebyshev–Sobolev consistently has the best mean accuracy, reaching around 97.5-98\% around degree 12 and maintaining this level. Chebyshev and Legendre–Sobolev also perform strongly, with mean accuracy peaking close to 97\% across higher degrees. Legendre lags slightly behind, plateauing closer to 96\%. These results indicate that both the choice of basis and the degree of the polynomial affect the recognition rate, with Sobolev-type and Chebyshev-based bases offering a modest but consistent advantage.

\subsection{Approximation of Handwritten symbol}

This experiment evaluates the effectiveness of different polynomial bases for handwriting representation. We plot a representative symbol from the ORCCA dataset and approximate it using Legendre, Chebyshev, Legendre-Sobolev, and Chebyshev-Sobolev polynomial bases. The approximated symbol is shown for degrees 5, 10, 15, and 20 in figure \ref{fig:trace-all}, with the original digital-ink trace overlapped by the approximated symbol.

\spacetune{~\vfill}
Figures \ref{fig:trace5} and \ref{fig:trace10} show the approximated curves for lower degrees, 5 and 10. The rough outline of the symbol appears, but finer details are missed. The biggest mistakes occur in the curved parts and at the ends of the trace. Legendre and Chebyshev bases follow the overall shape but are still missing in some places, while the Sobolev versions smooth out the lines, which makes them lose details in tight curves.

\spacetune{\newpage}
At degree 15, the results (Figure \ref{fig:trace15}) look closer to the original. The differences between the four bases are minor, and just small mismatches remain in certain curved spots. Chebyshev-based results usually fit those curves better, while the Sobolev polynomials make the lines look smoother and reduce small wiggles. At this point, all bases give a good balance between accuracy and stability, although each basis has its own style of error.

\spacetune{\balance}
At degree 20, the approximations (Figure \ref{fig:trace20}) look almost identical to the original symbol. The loops, intersections, and endpoints are all drawn very accurately, and the slight differences that remain are barely noticeable. The main effect of the basis here is on smoothness. Sobolev bases give a slightly steadier curve, while Chebyshev and Legendre bases also keep the fine details. This shows that the choice of basis matters at all degrees.

\section{Conclusions}
In this work, we have analyzed the choice of different polynomial bases, Legendre, Chebyshev, Legendre--Sobolev, and Chebyshev--Sobolev, in handwriting recognition. We investigated the trade-offs between computational time and classification accuracy using parametric curve representations and orthogonal polynomial approximations.

In theoretical analysis, we have bounded Sobolev norms in terms of coefficient norms, showing that minor coefficient variations produce proportionally bounded changes in both the function and its derivative, an essential property for robust handwriting recognition. The experimental results confirm these insights: Sobolev bases maintain lower coefficient norms even at higher degrees and yield smoother growth with increasing degree, making it well-suited for complex curves and higher-order approximations.

Legendre and Chebyshev bases offer faster computation time but suffer from instability at higher degrees. Chebyshev--Sobolev basis achieves significantly better recognition accuracy, although at the cost of increased computation time. The Chebyshev--Sobolev basis had the highest recognition accuracy among all the polynomial bases examined, peaking at more than 97.5\%.

Our results suggest that Sobolev orthogonal bases, especially Chebyshev--Sobolev, balance accuracy and stability well. These are good choices with practical applications in education, accessibility tools, and mathematical handwriting recognition.

\section*{Acknowledgements}
SMW thanks David Stoutemyer for the suggestion to consider Chebyshev-Sobolev polynomial bases after seeing the results for Legendre-Sobolev bases.
RMC and SMW acknowledge the support of the Natural Sciences and Engineering Research Council of Canada.

\spacetune{\vfill}
\bibliographystyle{IEEEtran}
\IfFileExists{IfExistsUseBBL.tex}{%

}{%
\bibliography{generic.bib}
}
\end{document}